\documentclass[runningheads]{llncs}
\usepackage[colorlinks]{hyperref}
\usepackage{graphicx}
\usepackage{comment}
\usepackage{amsmath,amssymb} % define this before the line numbering.
\usepackage{color}

\usepackage{caption}
\usepackage{subcaption}
\usepackage{makecell}
\usepackage{booktabs}
\newcommand\ignore[1]{}

\usepackage[table]{xcolor}
\newcommand{\grayrowcolor}{\rowcolor[gray]{0.85}}
\newcommand{\bluecellcolor}{\cellcolor{blue!10}}
\newcommand{\Sec}[1]{\S\ref{#1}}
\newcommand*\rot{\rotatebox{90}}

\begin{document}
% \linenumbers
\pagestyle{headings}
\mainmatter
\def\ECCVSubNumber{4670}  % Insert your submission number here

\title{Virtual Multi-view Fusion for 3D Semantic Segmentation} % Replace with your title

\titlerunning{Virtual Multiview Fusion}
\author{Abhijit Kundu \and
Xiaoqi Yin \and
Alireza Fathi \and
David Ross \and \\
Brian Brewington \and 
Thomas Funkhouser \and
Caroline Pantofaru
}

\authorrunning{A. Kundu et al.}

\institute{Google Research}
%******************
\maketitle

\begin{abstract}
Semantic segmentation of 3D meshes is an important problem for 3D scene understanding. In this paper we revisit the classic multiview representation of 3D meshes and study several techniques that make them effective for 3D semantic segmentation of meshes. Given a 3D mesh reconstructed from RGBD sensors, our method effectively chooses different virtual views of the 3D mesh and renders multiple 2D channels for training an effective 2D semantic segmentation model. Features from multiple per view predictions are finally fused on 3D mesh vertices to predict mesh semantic segmentation labels. Using the large scale indoor 3D semantic segmentation benchmark of ScanNet, we show that our virtual views enable more effective training of 2D semantic segmentation networks than previous multiview approaches. When the 2D per pixel predictions are aggregated on 3D surfaces, our virtual multiview fusion method is able to achieve significantly better 3D semantic segmentation results compared to all prior multiview approaches and competitive with recent 3D convolution approaches.

\end{abstract}

\section{Introduction}

Semantic segmentation of 3D scenes is a fundamental problem in computer vision.  Given a 3D representation of a scene (e.g., a textured mesh of an indoor environment), the goal is to output a semantic label for every surface point.  The output could be used for semantic mapping, site monitoring, training autonomous navigation, and several other applications.

State-of-the-art (SOTA) methods for 3D semantic segmentation currently use 3D sparse voxel convolution operators for processing input data.   For example, MinkowskiNet \cite{choy20194d} and SparseConvNet \cite{graham20183d} each load the input data into a sparse 3D voxel grid and extract features with sparse 3D convolutions.   These ``place-centric'' methods are designed to recognize 3D patterns and thus work well for types of objects with distinctive 3D shapes (e.g., chairs), and not so well for others (e.g., wall pictures).   They also take a considerable amount of memory, which limits spatial resolutions and/or batch sizes.

Alternatively, when posed RGB-D images are available, several researchers have tried using 2D networks designed for processing photographic RGB images to predict dense features and/or semantic labels and then aggregate them on visible 3D surfaces \cite{hermans2014dense,vineet2015incremental}, and others project features onto visible surfaces and convolve them further in 3D \cite{dai20183dmv,semanticpaint,jaritz2019multi,lai2014unsupervised}.   Although these ``view-centric'' methods utilize massive image processing networks pretrained on large RGB image datasets, they do not achieve SOTA performance on standard 3D segmentation benchmarks due to the difficulties of occlusion, lighting variation, and camera pose misalignment in RGB-D scanning datasets.  None of the view-based methods is currently in the top half of the current leaderboard for the 3D Semantic Label Challenge of the ScanNet benchmark.

In this paper, we propose a new view-based approach to 3D semantic segmentation that overcomes the problems with previous methods. The key idea is to use synthetic images rendered from ``virtual views'' of the 3D scene rather than restricting processing to the original photographic images acquired by a physical camera.   This approach has several advantages that address the key problems encountered by previous view-centric method \cite{boulch2018snapnet,lawin2017deep}.  First, we select camera intrinsics for virtual views with unnaturally wide field-of-view to increase the context observed in each rendered image.   Second, we select virtual viewpoints at locations with small variation in distances/angles to scene surfaces, relatively few occlusions between objects, and large surface coverage redundancy.  Third, we render non-photorealistic images without view-dependent lighting efffects and occlusions by backfacing surfaces -- i.e., virtual views can look into a scene from behind the walls, floors, and ceilings to provide views with relatively large context and little occlusion.  Fourth, we aggregate pixel-wise predictions onto 3D surfaces according to exactly known camera parameters of virtual views, and thus do not encounter ``bleeding'' of semantic labels across occluding contours. Fifth, virtual views during training and inference can mimic multi-scale training and testing and avoid scale in-variance issues of 2D CNNs. We can generate as many virtual views as we want during both training and testing. During training, more virtual views provides robustness due to data augmentation.  During testing, more views provides robustness due to vote redundancy. Finally, the 2D segmentation model in our multiview fusion approach can benefit from large image pre-training data like ImageNet and COCO, which are unavailable for pure 3D convolution approaches.

We have investigated the idea of using virtual views for semantic segmentation of 3D surfaces using a variety of ablation studies. We find that the broader design space of view selection enabled by virtual cameras can significantly boost the performance of multiview fusion as it allows us to include physically impossible but useful views (e.g., behind walls). For example, using virtual views with original camera parameters improves 3D mIoU by 3.1\% compared with using original photograpic images, using additional normal and coordinates channels and higher field of view can further boost mIoU by 5.7\%, and an additional gain of 2.1\% can be achieved by carefully selecting virtual camera poses to best capture the 3D information in the scenes and optimize for training 2D CNNs.

Overall, our simple system is able to achieve state-of-the-art results on both 2D and 3D semantic labeling tasks in ScanNet Benchmark~\cite{dai2017scannet}, and is significantly better than the best performing previous multi-view methods and very competitive with recent 3D methods based on convolutions of 3D point sets and meshes. In addition, we show that our proposed approach consistently outperforms 3D convolution and real multi-view fusion approaches when there are fewer scenes for training. Finally, we show that similar performance can be obtained with significantly fewer views in the inference stage. For example, multi-view fusion with $\sim$12 virtual views per scene will outperform that with all $\sim$1700 original views per scene.

The rest of the paper is organized as follows. We introduce the research landscape and related work in \Sec{sec:related_work}. We describe the proposed virtual multiview fusion approach in detail in \Sec{sec:methods}-\Sec{sec:fusion}. Experiment results and ablation studies of our proposed approach are presented in \Sec{sec:exp}. Finally we conclude the paper with discussions of future directions in \Sec{sec:conclusion}.

% Since 3D semantic labels are hard to obtain, exploiting rich unsupervised 3D constraints like 3D consistency of semantic label from different views is important. However incorporating such consistency constraints as explicit loss functions during training often involves complex sampling of nearby views from same scene and exploit consistency between small set of frames to fit a mini-batch size. We showed that simply running our multiview fusion model in basic self-training~\cite{scudder1965probability, he2019revisiting, xie2019self} framework implicitly adds consistency constraints.
\section{Related Work}\label{sec:related_work}

There has been a large amount of previous work on semantic segmentation of 3D scenes.  The following reviews only the most related work.

\vspace*{2mm}\noindent{\bf Multi-view labeling.}  Motivated by the success of view-based methods for object classification \cite{su2015multi}, early work on semantic segmentation of RGB-D surface reconstructions relied on 2D networks trained to predict dense semantic labels for RGB images.  Pixel-wise semantic labels were backprojected and aggregated onto 3D reconstructed surfaces via weighted averaging \cite{hermans2014dense,vineet2015incremental}, CRFs \cite{mccormac2017semanticfusion}, Bayesian fusion \cite{ma2017multi,vineet2015incremental,zhang2019large}, or 3D convolutions \cite{dai20183dmv,jaritz2019multi,lai2014unsupervised}.  These methods performed multiview aggregation only for the originally captured RGB-D photographic images, which suffer from limited fields-of-view, restricted viewpoint ranges, view-dependent lighting effects, and misalignments with reconstructed surface geometry, all of which reduce semantic segmentation performance.   To overcome these problems, some recent work has proposed using synthetic images of real data in a multiview labeling pipeline \cite{boulch2018snapnet,lawin2017deep,guerry2017snapnet}, but they still use camera parameters typical of real images (e.g., small field of view), propose methods suitable only for outdoor environments (lidar point clouds of cities), and do not currently achieve state-of-the-art results.

\vspace*{2mm}\noindent{\bf 3D convolution.} Recent work on 3D semantic segmentation  has focused on methods that extract and classify features directly with 3D convolutions.  Network architectures have been proposed to extract features from 3D point clouds \cite{pham2019jsis3d,qi2017pointnet,qi2017pointnet++,shi2019pv,thomas2019kpconv,jsenet}, surface meshes \cite{hanocka2019meshcnn,huang2019texturenet}, voxel grids \cite{song2017semantic}, and octrees \cite{riegler2017octnet}.  Current state-of-the-art methods are based on sparse 3D voxel convolutions \cite{choy20194d,choy2019fully,graham20183d}, where submanifold sparse convolution operations are used to compute features on sparse voxel grids.   These methods utilize memory more efficiently than dense voxel grids, but are still limited in spatial resolution in comparison to 2D images and can train with supervision only on 3D datasets, which generally are very small in comparison to 2D image datasets.

\vspace*{2mm}\noindent{\bf Synthetic data.}  Other work has investigated training 2D semantic segmentation networks using computer graphics renderings of 3D synthetic data \cite{zhang2017physically}.   The main advantage of this approach is that image datasets can be created with unlimited size by rendering novel views of a 3D scene \cite{li2018interiornet,mccormac2017scenenet}.   However, the challenge is generally domain adaptation -- networks trained on synthetic data and tested on real data usually do not perform well.  Our method avoids this problem by training and testing on synthetic images rendered with the same process.

% Source image at: https://docs.google.com/presentation/d/1gRueRg7yGnr8acuTjCvfpMRla7SjVhLNMqiRl5QE3kQ/edit#slide=id.g704128e989_0_201
\begin{figure}
    \centering
    \includegraphics[width=0.95\linewidth]{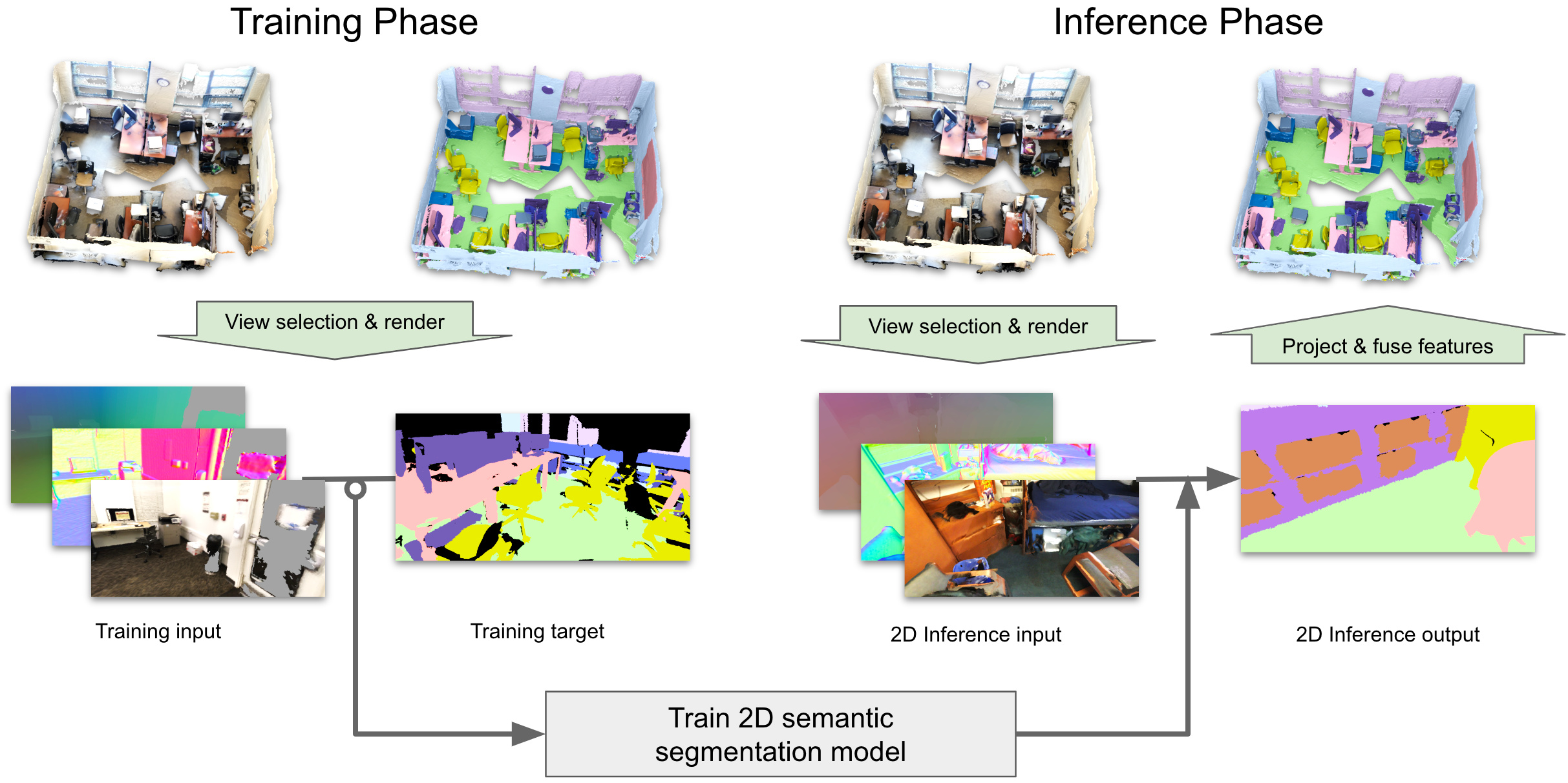}
    \caption{Virtual multi-view fusion system overview.}
    \label{fig:overview}
\end{figure}

% Source image at: https://docs.google.com/presentation/d/1gRueRg7yGnr8acuTjCvfpMRla7SjVhLNMqiRl5QE3kQ/edit#slide=id.g704128e989_0_102
\begin{figure}[t]
    \centering
    \includegraphics[width=0.95\linewidth]{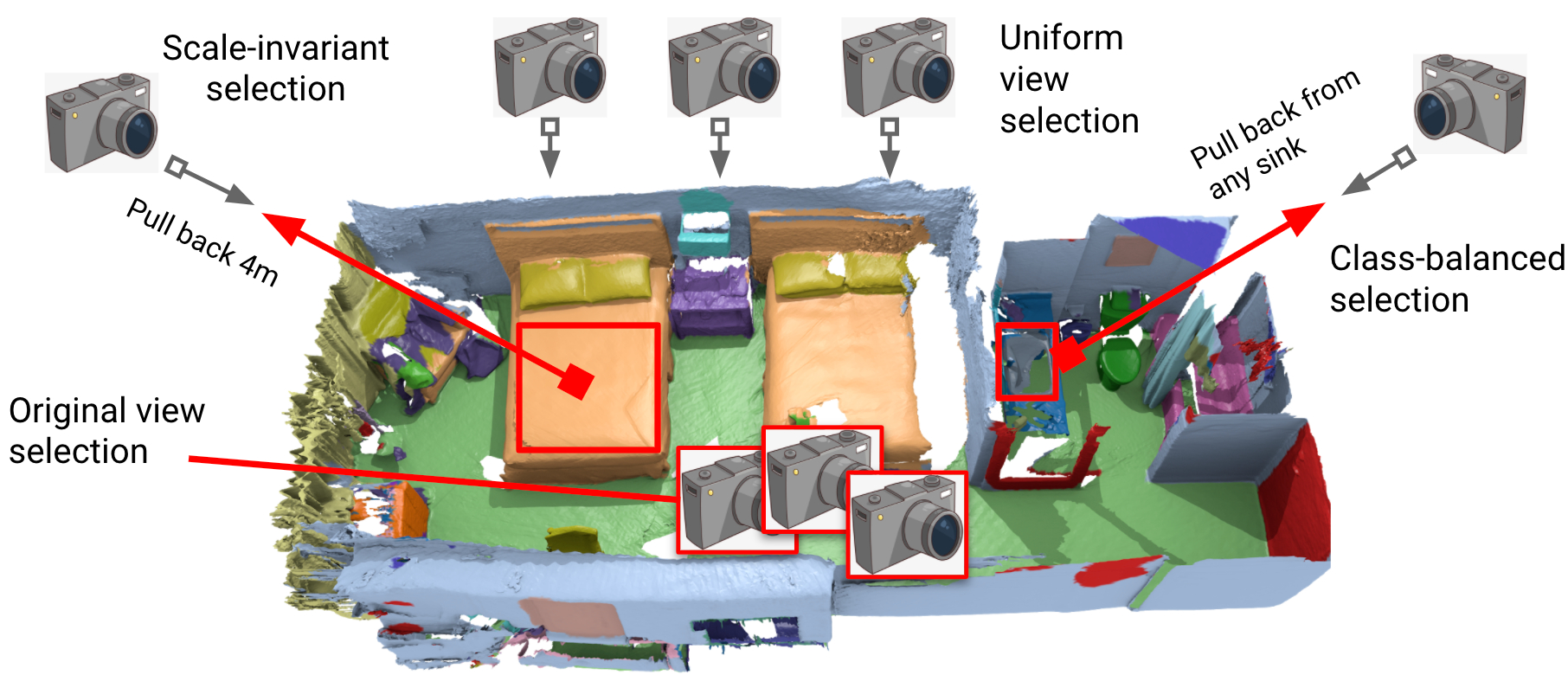}
    \caption{Proposed virtual view selection approaches.}
    \label{fig:view_selection}
\end{figure}
\vspace{-5mm}

\section{Method Overview} \label{sec:methods}

% Talk about rendering virtual views.
% Talk about training a 2D model.
% 3D fusion.

The proposed multiview fusion approach is illustrated in Figure~\ref{fig:overview}. At a high level, it consists of the following steps.

\vspace*{2mm}\noindent{\bf Training stage.} During the training stage, we first select virtual views for each 3D scene, where for each virtual view we select camera intrinsics, camera extrinsics, which channels to render, and rendering parameters (e.g., depth range, backface culling). We then generate training data by rendering the selected virtual views for the selected channels and ground truth semantic labels. We train 2D semantic segmentation models using the rendered training data and use the model in the inference stage.

\vspace*{2mm}\noindent{\bf Inference stage.} At inference stage, we select and render virtual views using a similar approach as in the training stage, but without the ground truth semantic labels. We conduct 2D semantic segmentation on the rendered virtual views using the trained model, project the 2D semantic features to 3D, then derive the semantic category in 3D by fusing multiple projected 2D semantic features.

\section{Virtual view selection}

Virtual view selection is central to the proposed multiview fusion approach as it brings key advantages over multiview fusion with original image views. First, it allows us to freely select camera parameters that work best for 2D semantic segmentation tasks, and with any set of 2D data augmentation approaches. Second, it significantly broadens the set of views to choose from by relaxing the physical constraints of real cameras and allowing views from unrealistic but useful camera positions that significantly boost model performance, e.g. behind a wall. Third, it allows 2D views to capture additional channels that are difficult to capture with real cameras, e.g., normals and coordinates. Finally, by selecting and rendering virtual views, we have essentially eliminated any errors in the camera calibration and pose estimation, which are common in the 3D reconstruction process. Lastly, sampling views consistently at different scales resolves scale in-variance issues of traditional 2D CNNs.

\begin{figure}[t]
    \centering
    \includegraphics[width=0.95\linewidth]{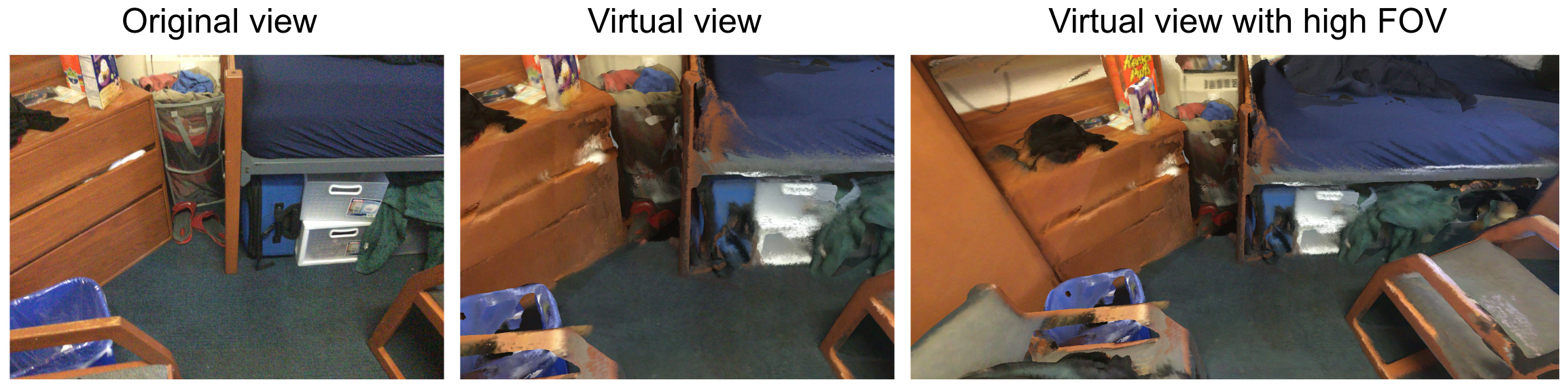}
    \caption{Original views vs. virtual views. High FOV provides larger context of the scene which helps 2D perception, e.g., the chair in the bottom right corner is partially represented in the original view but can easily segmented in the high FOV virtual view.}
    \label{fig:virtual_view}
\end{figure}

\vspace*{2mm}\noindent{\bf Camera intrinsics.} A significant constraint of original image views is the FOV - images may have been taken very close to objects or walls, say, and lack the object features and context necessary for accurate classification. Instead, we use a pinhole camera model with significantly higher field of view (FOV) than the original cameras, providing larger context that leads to more accurate 2D semantic segmentation \cite{Mottaghi_2014_CVPR}. Figure~\ref{fig:virtual_view} shows an example of original views compared with virtual views with high FOV.

\begin{figure}[t]
    \centering
    \includegraphics[width=0.95\linewidth]{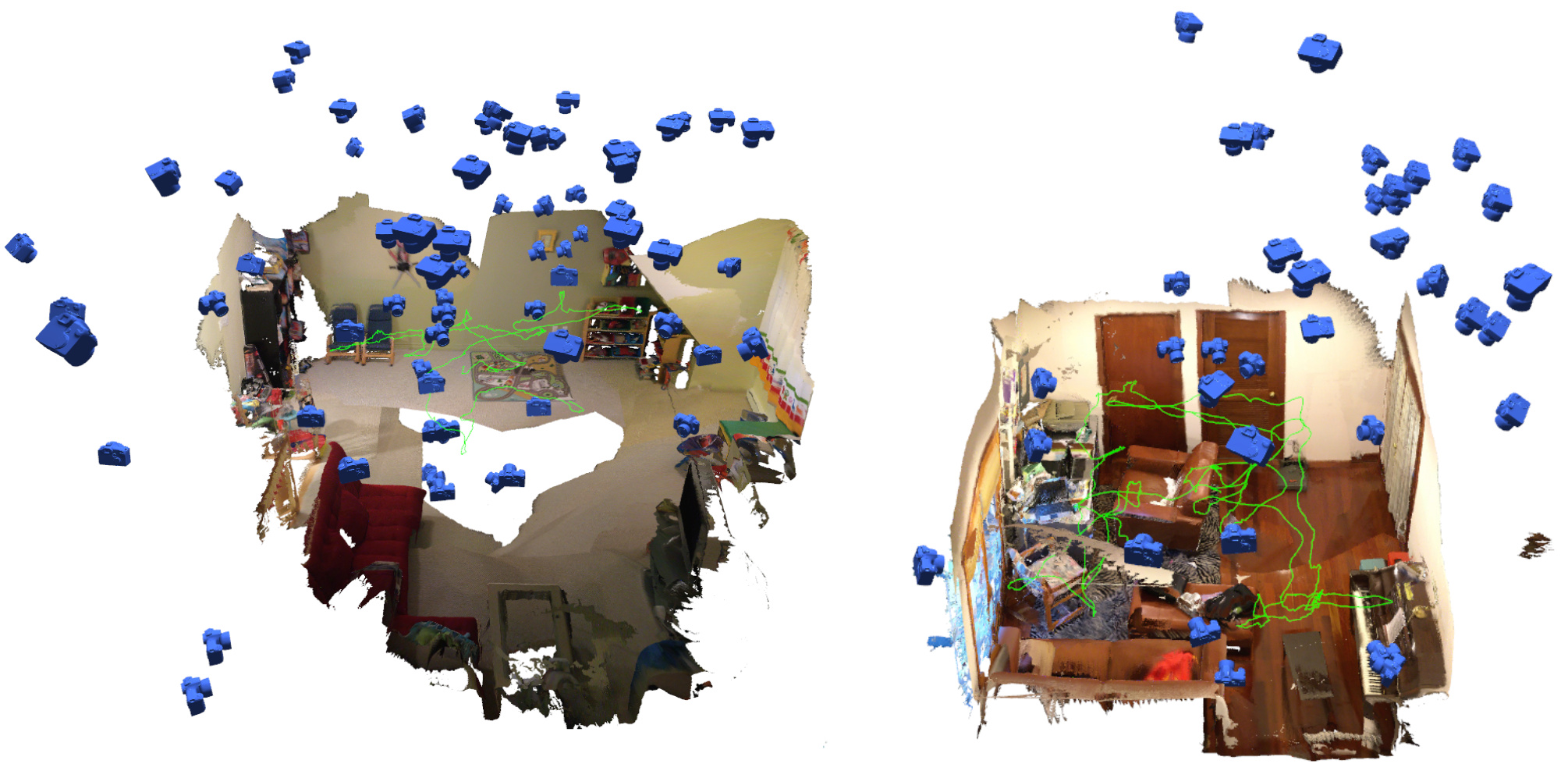}
    \caption{Example virtual view selection on two ScanNet scenes. Green curve is the trajectory of the original camera poses; Blue cameras are the selected views with proposed approaches. Note that we only show a random subset of all selected views for illustration purposes.}
    \label{fig:virtual_extrinsics}
\end{figure}

\vspace*{2mm}\noindent{\bf Camera extrinsics.} We use a mixture of the following sampling strategies to select camera extrinsics as shown in Figure~\ref{fig:view_selection} and Figure~\ref{fig:virtual_extrinsics}.
\begin{itemize}
    \item Uniform sampling. We want to uniformly sample camera extrinsics to generate many novel views,  independently from the specific structure of the 3D scene. Specifically, we use top-down views from uniformly sampled positions at the top of the 3D scene, as well as views that look through the center of the scene but with uniformly sampled positions in the 3D scene.
    \vspace*{1mm}\item Scale-invariant sampling. As 2D convolutional neural networks are generally not scale invariant, the model performance may suffer if the scales of views do not match the 3D scene. To overcome this limitation, we propose sampling views at a range of scales with respect to segments in the 3D scene. Specifically, we do an over-segmentation of the 3D scene, and for each segment, we position the cameras to look at the segment by pulling back to a certain range of distances along the normal direction. We do a depth check to avoid occlusions by foreground objects. If backface culling is disabled in the rendering stage (discussed in more detail below), we do a ray tracing and drop any views blocked by the backfaces. Note the over-segmentation of the 3D scene is unsupervised and does not use the ground truth semantic labels, so the scale-invariant sampling can be applied both in the training and inference stages.
    \vspace*{1mm}\item Class-balanced sampling. Class balancing has been extensively used as data augmentation approaches for 2D semantic segmentation. We conduct class balancing by selecting views that look at mesh segments of under-represented semantic categories, similar to the scale-invariant sampling approach. Note this sampling approach only applies to the training stage when the ground truth semantic labels are available.
    \vspace*{1mm}\item Original views sampling. We also sample from the original camera views as they represent how a human would choose camera views in the real 3D scene with real physical constraints. Also, the 3D scene is reconstructed from the original views, so including them can make sure we cover corner cases that would otherwise be difficult as random virtual views. 
\end{itemize}

\begin{figure}[t]
    \centering
    \includegraphics[width=0.85\linewidth]{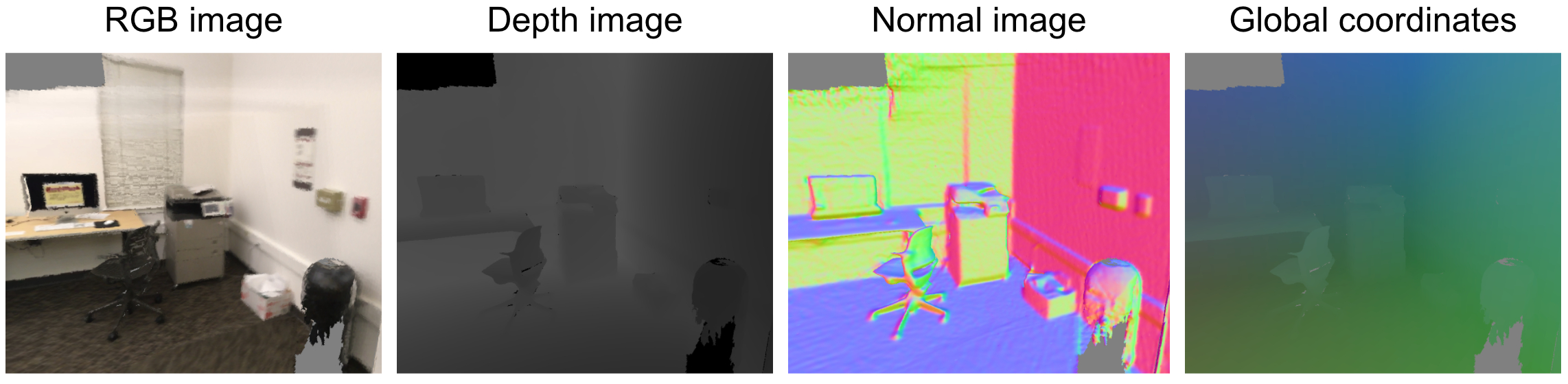}
    \caption{Example virtual rendering of selected channels.}
    \label{fig:rendered_channels}
\end{figure}

\vspace*{2mm}\noindent{\bf Channels for rendering.} To exploit all the 3D information available in the scene, we render the following channels: RGB color, normal, normalized global XYZ coordinates. The additional channels allow us to go beyond the limitations of the existing RGB-D sensors. While depth image also contains the same information, we think normalized global coordinate image makes the learning problem simpler as now just like the normal and color channel, coordinate values of the same 3D point is view invariant. Figure~\ref{fig:rendered_channels} shows example rendered views of the selected channels.

\begin{figure}[t]
    \centering
    \includegraphics[width=0.75\linewidth]{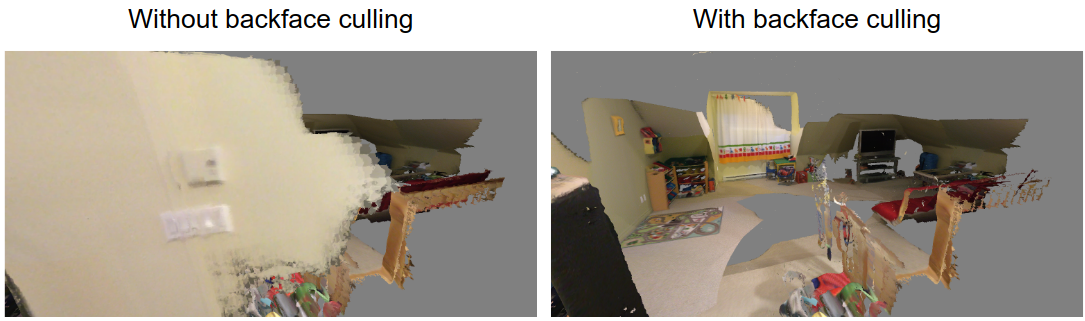}
    \caption{Effect of backface culling. Backface culling allows the virtual camera to see more context from views that are not physically possible with real cameras.}
    \label{fig:backface_culling}
\end{figure}

\vspace*{2mm}\noindent{\bf Rendering parameters.} We turn on backface culling in the rendering so that the backfaces do not block the camera views, further relaxing the physical constraints of the 3D scene and expanding the design space of the view selection. For example, as shown in Figure~\ref{fig:backface_culling}, in an indoor scenario, we can select views from outside a room which typically include more context of the room and can potentially improve model performance; On the other hand, with backface culling turned off, we either are constrained ourselves to views inside the room therefore limited context, or suffer from high occlusion by the backfaces of the walls.

\vspace*{2mm}\noindent{\bf Training vs. inference stage.} We want to use similar view selection approaches for the training and inference stages to avoid creating a domain gap, e.g., if we sampled many top-down views in the training stage but used lots of horizontal views in the inference stage. The main difference between the view selection strategies between the two stages is the class-balancing which can only be done in the training stage. Also, while the inference cost may matter in real-world applications, in this paper we consider offline 3D segmentation tasks and do not optimize the computation cost in either stage, so we can use as many virtual views as needed in either stage.

\section{Multiview Fusion}\label{sec:fusion}

\subsection{2D semantic segmentation model}

% Class balancing.
With rendered virtual views as training data, we are now ready to train a 2D semantic segmentation models. We use a xcpetion65~\cite{chollet2017xception} feature extractor and DeeplabV3+~\cite{deeplabv3plus2018} decoder. We initialize our model from pre-trained classification model checkpoints trained on ImageNet. When training a model with additional input channels like normal image and co-ordinate image we modify the first layer of the pre-training checkpoints by tiling the weights across the additional channels and normalize them across each spatial position such that the sum of weights along the channel dimension remains the same.

\subsection{3D fusion of 2D semantic features}

During inference, we run the 2D semantic segmentation model on virtual views and obtain image features (e.g., unary probabilities for each pixel). To project the 2D image features to 3D, we use the following approach: We render a depth channel on the virtual views; For each 3D point, we project it back to each of the virtual views, and accumulate the image feature of the projected pixel only if the depth of the pixel matches the point-to-camera distance. This approach achieves better computational efficiency than the alternative approach of casting rays from each pixel to find the 3D point to aggregate. First, the number of 3D points in a scene are much less than the total number of pixels in all rendered images of the scene. Secondly, projecting a 3D point with a depth check is faster than operations involving ray casting.

Formally, let $\mathbf{X}_k\in \mathbb{R}^3$ be the 3D position of the $k$th point, $\mathbf{x}_{k,i}\in \mathbb{R}^2$ be the pixel coordinates by projecting the $k$th 3D point to virtual view $i\in\mathcal{I}$, $\mathbf{K}_i$ be its instrinsics matrix while $\mathbf{R}_i$ be the rotation, $\mathbf{t}_i$ the translation in the extrinsics, $\mathcal{A}_i$ be the set of valid pixel coordinates.  Let $c_{k,i}$ be the distance between the position of camera $i$ and $k$th 3D point. We have:
\begin{equation}
    \mathbf{x}_{k,i} = \mathbf{K}_i(\mathbf{R}_i\mathbf{X}_k + \mathbf{t}_i)
\end{equation}
\begin{equation}
    c_{k, i} = \left\lVert \mathbf{X}_k - \mathbf{R}_i^{-1}\mathbf{t}_i\right\rVert_2
\end{equation}

Let $\mathcal{F}_k$ be the set of image features projected to the $k$th 3D point, $\mathbf{f}_i(\cdot)$ be the mapping from pixel coordinates in virtual image $i$ to the image feature vector, $d_i(\cdot)$ be the mapping from pixel coordinates to the depth since we render depth channel. Then:
\begin{equation}
    \mathcal{F}_k = \left\{ \mathbf{f}_i(\mathbf{x}_{k,i}) \ \Large{\vert} \ \mathbf{x}_{k,i} \in \mathcal{A}_i, \left\lvert d_i(\mathbf{x}_{k,i}) - c_{k,i} \right\rvert  < \delta, \forall i\in\mathcal{I} \right\}
\end{equation}
where $\delta>0$ is the threshold for depth matching.

To fuse projected features $\mathcal{F}_k$ for 3D point $k$, we simply take the average of all features in $\mathcal{F}_k$ and obtain the fused feature. There simple fusion function was better than other alternatives like picking the category with maximum probability across all projected features.

% Equation for 3D fusion.
% It is important to iterate over 3D points rather projection each image pixel.

% \subsection{Self-training for implicit Consistency}

% \begin{figure}[t]
%     \centering
%     \includegraphics[width=0.8\linewidth]{figures/self_training.png}
%     \caption{Multiview fusion with Self-training.}
%     \label{fig:self_training}
%     \vspace{-0.2cm}
% \end{figure}

% By fusing semantic features observed from multiple views, many prediction errors and inconsistencies from single views can be significantly reduced. As discussed in Section~\ref{sec:exp}, we can get significant performance boost on the 2D image IoU at inference stage by conducting multiview fusion then projecting the 3D fused features back to the original 2D views. We can then add the reprojected views with ``pseudo" ground truth semantic labels to the training set, and retrain the 2D semantic segmentation models. Figure~\ref{fig:self_training} shows a high-level overview of the proposed self training approach. This process can be iterated for multiple times.

\section{Experiments}\label{sec:exp}

We ran a series of experiments to evaluate how well our proposed method for 3D semantic segmentation of RGB-D scans works compared to alternative approaches and to study how each component of our algorithm affects the results.

\subsection{Evaluation on ScanNet dataset.}

\begin{table}
\centering
\begin{tabular}{|l|c|c|c|}
\hline
\grayrowcolor Method& 3D mIoU (val split) & 3D mIoU (test split) & 2D mIoU (test split)  \\ \hline \hline
PointNet~\cite{qi2017pointnet}              & 53.5   &55.7    &-\\
3DMV~\cite{dai20183dmv}                     & -      &48.4    &49.8\\
SparseConvNet~\cite{graham20183d}           & 69.3   &72.5    &-\\
PanopticFusion~\cite{narita2019panopticfusion} & -      &52.9    &-\\
PointConv~\cite{wu2019pointconv}            & 61.0   &66.6    &-\\
JointPointBased~\cite{jointpointbased}      & 69.2   &63.4    &-\\
SSMA~\cite{valada2019self}                  & -      &-       &57.7\\
KPConv~\cite{thomas2019kpconv}              & 69.2   &68.4    &-\\
MinkowskiNet~\cite{choy20194d}              & 72.2   &73.6.   &-\\
PointASNL~\cite{yan2020pointasnl}           & 63.5   &66.6    &-\\
OccuSeg~\cite{han2020occuseg}               & -   &\textbf{76.4}    &-\\
JSENet~\cite{jsenet}                        & -      &69.9    &-\\
\hline
Ours                                        & \textbf{76.4}   &74.6 &\textbf{74.5}\\
\hline
\end{tabular}
\vspace{2mm}
\caption{Semantic segmentation results on ScanNet validation and test splits.}
\label{tab:scannet_table}
\end{table}

\begin{figure}[tb]
\includegraphics[width=1.0\linewidth]{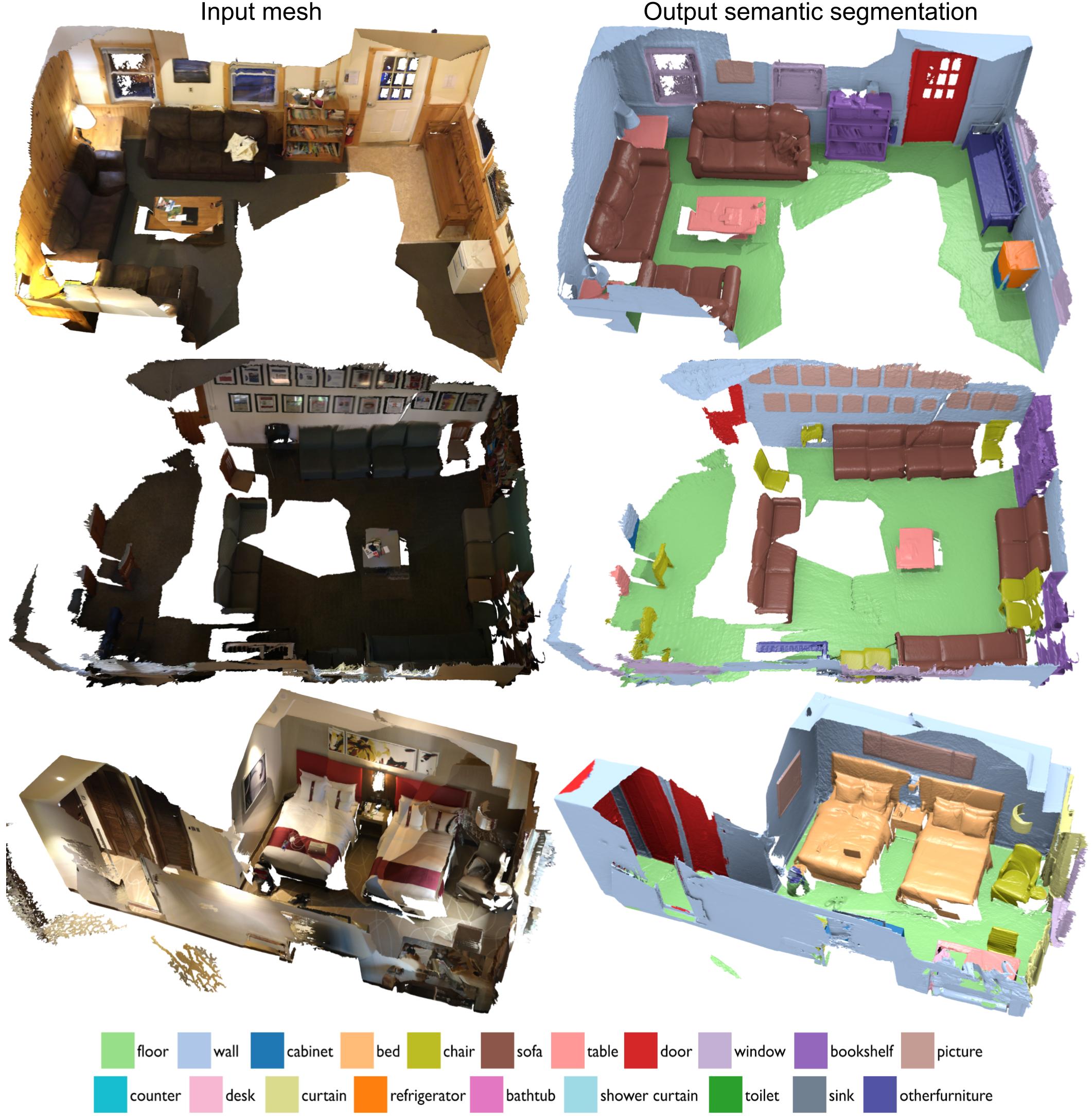}
\caption{Qualitiative 3D semantic segmentation results on ScanNet test set.}
\label{fig:qualitative_scannet}
\end{figure}

We evaluate our approach on ScanNet dataset~\cite{dai2017scannet}, on the hidden test set for the task of both 3D mesh semantic segmentation and 2D image semantic segmentation. We also perform a detailed ablation study on the validation set of ScanNet in \Sec{sec:experiments_ablation}. Unlike our ablation studies, we use xception101~\cite{chollet2017xception} as the 2D backbone and we additionally use ADE20K~\cite{zhou2017scene} for pre-training the 2D segmentation model.  We compare our virtual mutiview-fusion approach against state-of-the-art methods for 3D semantic segmentation, most of which utilize 3D convolutions of sparse voxels or point clouds. We also compare our 2D image segmentation results obtained by projecting back 3D labels obtained by our multiview fusion approach. Results are available in Table \ref{tab:scannet_table}.

From these results, we see that our approach outperforms previous approaches based on convolutions of 3D point sets~\cite{qi2017pointnet,wu2019pointconv,thomas2019kpconv,yan2020pointasnl,jsenet}, and it achieves results comparable to the SOTA methods based on sparse voxel convolutions~\cite{graham20183d,choy20194d,han2020occuseg}. Our method achieves the best 2D segmentation results (74.5\%). In \Sec{sec:experiments_ablation}, we also demonstrate improvement in single frame 2D semantic segmentation.

Our approach performs significantly better than any previous multiview fusion methods~\cite{dai20183dmv,narita2019panopticfusion} on ScanNet semantic labeling benchmark. The mean IoU of the previously best performing multiview method on the ScanNet test set is 52.9\% \cite{narita2019panopticfusion}, which is significantly less than our results of 74.6\%.  By using our virtual views, we are able to learn 2D semantic segmentation networks that provide more accurate and more consistent semantic labels when aggregated on 3D surfaces.  The result is semantic segmentations of high accuracy and sharp boundaries, as shown in Figure~\ref{fig:qualitative_scannet}.

\begin{table*}
\resizebox{\textwidth}{!}{%
\begin{tabular}{|c|c|ccccccccccccc|}
\hline
\grayrowcolor Method & mIOU & ceiling & floor & wall & beam & column & window & door & chair & table  & bookcase & sofa & board&  clutter  \\ \hline \hline
PointNet\cite{qi2017pointnet} &41.09 &88.8 &97.3 &69.8& 0.1 &3.9 &46.3 &10.8 &52.6 &58.9 &40.3 &5.9 &26.4 &33.2 \\
SegCloud\cite{segcloud} &48.92 &90.1 &96.1 &69.9 &0.0 &18.4 &38.4 &23.1 &75.9 &70.4 &58.4 &40.9 &13.0 &41.6 \\
TangentConv\cite{tangentconv} &52.80 &90.5 &97.7 &74.0 &0.0 &20.7 &39.0 &31.3 &77.5 &69.4 &57.3 &38.5 &48.8 &39.8 \\
3D RNN\cite{3drnn} &53.40 &95.2& 98.6 &77.4 &0.8 &9.8 &52.7 &27.9 &76.8 &78.3 &58.6 &27.4 &39.1 &51.0 \\
PointCNN\cite{pointcnn} &57.26 &92.3 &98.2 &79.4 &0.0 &17.6 &22.7 &62.1 &80.6 &74.4 &66.7 &31.7 &62.1 &56.7 \\
SuperpointGraph\cite{superpointgraph} & 58.04 & 89.4 &96.9 &78.1& 0.0 &42.8& 48.9 &61.6 &84.7 &75.4 &69.8 &52.6 &2.1 &52.2 \\
PCCN\cite{pccn} & 58.27 &90.3 &96.2 &75.9 &0.3 &6.0 &69.5 &63.5 &66.9 &65.6 &47.3 &68.9 &59.1 &46.2 \\
PointASNL\cite{yan2020pointasnl} &62.60 &94.3 &98.4 &79.1 & 0.0 &26.7 &55.2 &66.2 &83.3 &86.8 &47.6 &68.3 &56.4 &52.1 \\
MinkowskiNet\cite{choy20194d} & 65.35 & 91.8 & 98.7 & 86.2 & 0.0&  34.1&  48.9 & 62.4 & 89.8 & 81.6 & 74.9 & 47.2 & 74.4 & 58.6 \\
\textbf{Ours} & \textbf{65.38}& 92.9 &  96.9& 85.5& 0.8 & 23.3 & 65.1 & 45.7 & 85.8 & 76.9 & 74.6 & 63.1 & 82.1 & 57.0 \\
\hline
\end{tabular}}
\vspace{2mm}
\caption{Results on the Stanford 3D Indoor Spaces (S3DIS) dataset~\cite{s3dis_arxiv}. Following previous works we use Fold-1 split with Area5 as the test set.}
\label{tab:s3dis}
\end{table*}

\begin{figure}[ht]
\centering
\includegraphics[width=1.0\linewidth]{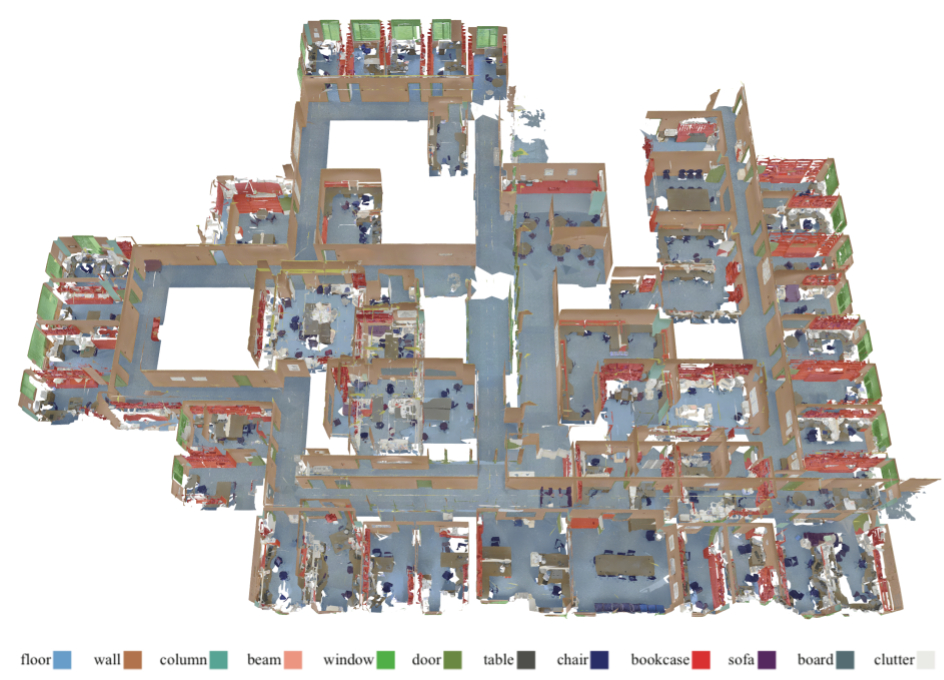}
\caption{Qualitiative 3D semantic segmentation results on Area5 of Stanford 3D Indoor Spaces (S3DIS) Dataset. Semantic label colors are overlayed on the textured mesh. \textit{Ceiling} not shown for clarity. }
\label{fig:qualitative_s3dis}
\vspace{-0.6cm}
\end{figure}

\subsection{Evaluation on Stanford 3D Indoor Spaces (S3DIS).}

We also evaluated our method on the Stanford Large-Scale 3D Indoor Spaces Dataset (S3DIS)~\cite{s3dis_arxiv,s3dis_cvpr16} for the task of semantic 3D segmentation.  The proposed virtual multi-view fusion approach achieves 65.4\% 3D mIoU, outperforming recent SOTA methods MinkowskiNet~\cite{choy20194d} (65.35\%) and PointASNL~\cite{yan2020pointasnl} (62.60\%). See Table~\ref{tab:scannet_table} for quantitative evaluation. Figure~\ref{fig:qualitative_s3dis} shows the output of our approach on Area5 scene from S3DIS dataset.

\subsection{Ablation Studies}\label{sec:experiments_ablation}

We investigate which aspects of our proposed method make the most difference we performed ablation study on the ScanNet~\cite{dai2017scannet}. To perform this experiment, we started with a baseline method that trains a model to compute 2D semantic segmentation for the original photographic images, uses it to predict semantics for all the original views in the validation set, and then aggregates the class probabilities on backprojected 3D surfaces using the simple averaging method described in Section~\ref{sec:methods}.   This mean class IoU of this baseline result is shown in the top row of Table~\ref{tab:view_ablation}.   We then performed a series of tests where we included features of our virtual view algorithm one-by-one and measured the impact on performance.   The second row shows the impact of using rendered images rather than photographic ones; the third shows the impact of adding additional image normal and coordinate channels captured during rendering; the fourth row shows the impact of rendering images with two times larger field-of-view; the fifth row shows the impact of our virtual viewpoints selection algorithm. We find that each of these ideas improves the 3D segmentation IoU performance significantly.  

Specifically, with fixed camera extrinsics matching the original views, we compare the effect of virtual view renderings versus the original photographic images: using virtual views leads to 3.1\% increase of 3D mIoU as it removes any potential errors in the 3D reconstruction and pose estimation process.  Using additional channels of normal and global coordinates achieves another 2.9\% performance boost in 3D mIoU as it allows the 2D semantic segmentation model to exploit the 3D information in the scene other than RGB.  Increasing the FOV further improves the 3D mIoU by 1.8\% since it allows the 2D model to use more context.  Lastly, view sampling with backface culling achieves the best performance and an 2.2\% improvement compared to the original views, showing that the camera poses can significantly affect the perception of 3D scenes. % Similar performance improvements are observed in the 2D mIoU on the original images. 
In addition, we compute and compare a) the \textit{single-view} 2D image mIoU,  which compares 2D ground truth with the prediction of a 2D semantic segmentation model from single image, and b) \textit{multi-view} 2D image mIoU, which compares ground truth with the reprojected semantic labels from the 3D semantic segmentation after multiview fusion. In all cases, we observed consistent improvements of 2D image mIoU after multiview fusion by a margin of 5.3\% to 8.4\%. This shows the multiview fusion effectively aggregates the observations and resolves the inconsistency between different views. Note that the largest single-view to multi-view improvement (8.4\%) is observed in the first row, i.e., on the original views, which confirms our hypothesis of potential errors and inconsistency in the 3D reconstruction and pose estimation process and the advantage of virtual views on removing these inconsistencies.

\begin{table*}
\resizebox{\textwidth}{!}{%
\begin{tabular}{|c|c|c|c|c|c|c|}
\hline
\grayrowcolor Image Type & Input Image Channels & Intrinsics & Extrinsics & \makecell{2D Image IoU\\(Single View)} & \makecell{3D Mesh IoU\\(Multiview)} & \makecell{2D Image IoU\\(Multiview)} \\ \hline \hline
Real & RGB & Original & Original & 60.1 & 60.1 & 68.5 \\ 
\bluecellcolor Virtual & RGB & Original & Original & 64.4 & 63.2 & 69.8 \\ 
\bluecellcolor Virtual & \bluecellcolor RGB + Normal + Coordinates & Original & Original & 66.1 & 66.1 & 70.8 \\ 
\bluecellcolor Virtual & \bluecellcolor RGB + Normal + Coordinates & \bluecellcolor High FOV & Original & 66.9 & 67.9  & 72.2 \\
\bluecellcolor Virtual & \bluecellcolor RGB + Normal + Coordinates & \bluecellcolor High FOV & \bluecellcolor View sampling & 67.0 & 70.1 & 74.9 \\  \hline
\end{tabular}}
\vspace{2mm}
\caption{\textbf{Evaluation on 2D and 3D Semantic segmentation tasks on ScanNet validation set.}  Ablation study evaluating the impact of sequentially adding features from our proposed virtual view fusion algorithm. The top row shows results of the traditional semantic segmentation approach with multiview fusion -- where all semantic predictions are made on the original captured input images.   Subsequent rows show the impact of gradually replacing characteristics of the original views with virtual ones. The bottom row shows the performance of our overall method using virtual views.}
\label{tab:view_ablation}
\vspace{-0.6cm}
\end{table*}

\subsubsection{Effect of Training Set Size}

\begin{figure}[bt]
\centering
  \begin{subfigure}[b]{0.48\textwidth}
    \includegraphics[width=\linewidth]{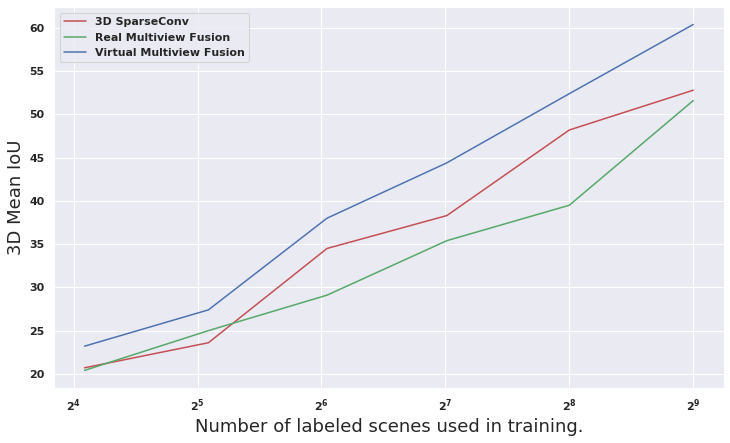}
    \caption{Effect of training data size on 3D segmentation IoU. Virtual multiview fusion model gives the better performance even when training data is small. Our hypothesis is that virtual view provides better data augmentation than simple 2D image level augmentations. Data augmentation is important with less training data}
    \label{fig:train_size}
  \end{subfigure}
~
  \begin{subfigure}[b]{0.48\textwidth}
    \includegraphics[width=\linewidth]{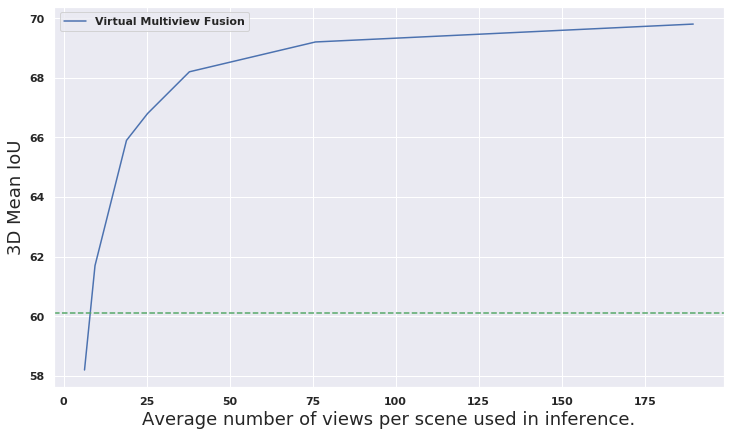}
    \caption{Effect of number of views used at inference time on 3D segmentation. The dotted green line shows the best mIoU ($60.1$) obtained with multi-view fusion using all original views ($\approx$ 1700 views per scene). Our virtual multiview fusion model achieves the same accuracy with just $\approx$ 12 views per scene.}
    \label{fig:inference_size}
  \end{subfigure}
  \caption{Impact of data size (number of views) during training and inference.}
\end{figure}

Our next experiment investigates the impact of the training set size on our algorithm.  We hypothesize that generating large numbers of virtual views provides a form of data augmentation that improves generalizability of small training sets.   To test this idea, we randomly sampled different numbers of scenes from the training set and trained our algorithm only on them.   We compare performance of multiview fusion using a 2D model trained from virtual views rendered from those scenes versus from the original photographic images, as well as a 3D convolution method SparseConv (Figure~\ref{fig:train_size}). Note that we conduct the experiments on ScanNet low resolution meshes while for others we use high resolution ones. For virtual/real multiview fusion approaches, we use the same set of views for each scene across different experiments.  We find that the virtual multiview fusion approach consistently outperforms 3D SparseConv and real multiview fusion even with a small number of scenes.

\subsubsection{Effect of number of views at Inference}

Next we investigate the impact of number of virtual views used in the inference stage on our algorithm. We run our virtual view selection algorithms on the ScanNet validation dataset, run a 2D model on them, and then do multiview fusion using only a random subset of the virtual views. As shown in Figure~\ref{fig:inference_size}, the 3D mIoU increases with the number of virtual views with diminishing returns. The virtual multiview fusion approach is able to achieve good performance even with a significantly smaller inference set. For example, while we achieve 70.1\% 3D mIoU with all virtual views ($\sim$2000 views per scene), we can reach 61.7\% mIoU even with $\sim$10 views per scene, and 68.2\% with $\sim$40 views per scene. In addition, the result shows that using more views selected with the same approach as for training views does not negatively affect the multiview fusion performance, which is not obvious as the confident but wrong prediction of one single view can harm the overall performance.

\section{Conclusion}\label{sec:conclusion}

In this paper, we propose a virtual multiview fusion approach to 3D semantic segmentation of textured meshes.   This approach builds off a long history of representing and labeling meshes with images, but introduces several new ideas that significantly improve labeling performance: virtual views with additional channels, back-face culling, wide field-of-view, multiscale aware view sampling.  As a result, it overcomes the 2D-3D misalignment, occlusion, narrow view, and scale invariance issues that have vexed most previous multiview fusion approaches.

The surprising conclusion from this paper is that multiview fusion algorithms are a viable alternative to 3D convolution for semantic segmentation of 3D textured meshes.  Although early work on this task considered multiview fusion, the general approach has been abandoned in recent years in favor of 3D convolutions of point clouds and sparse voxel grids. This paper shows that the simple method of carefully selecting and rendering virtual views enables multiview fusion to outperform almost all recent 3D convolution networks. It is also complementary to more recent 3D approaches. We believe this will encourage more researchers to build on top of this.

\clearpage
\bibliographystyle{splncs04}
\bibliography{references}

\section{Appendix}\label{sec:appendix}

\begin{figure}
\includegraphics[width=\linewidth]{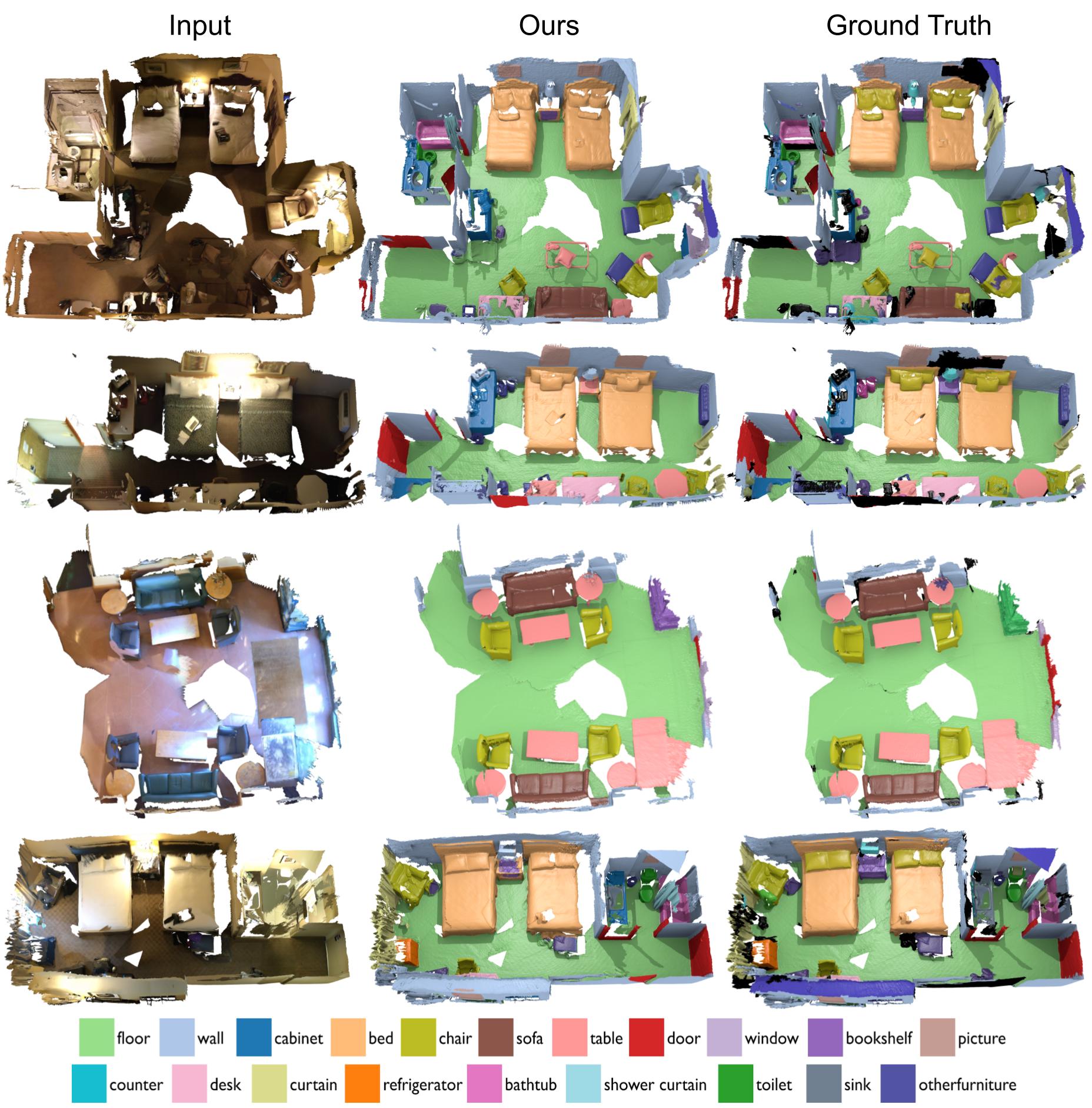}
\caption{Qualitative 3D semantic segmentation results on ScanNet validation set for scenes \texttt{scene0645\_00}, \texttt{scene0221\_00}, \texttt{scene0549\_00}, and \texttt{scene0435\_01} respectively.}
\label{fig:qualitative_scannet_val1}
\end{figure}

\begin{figure}
\includegraphics[width=\linewidth]{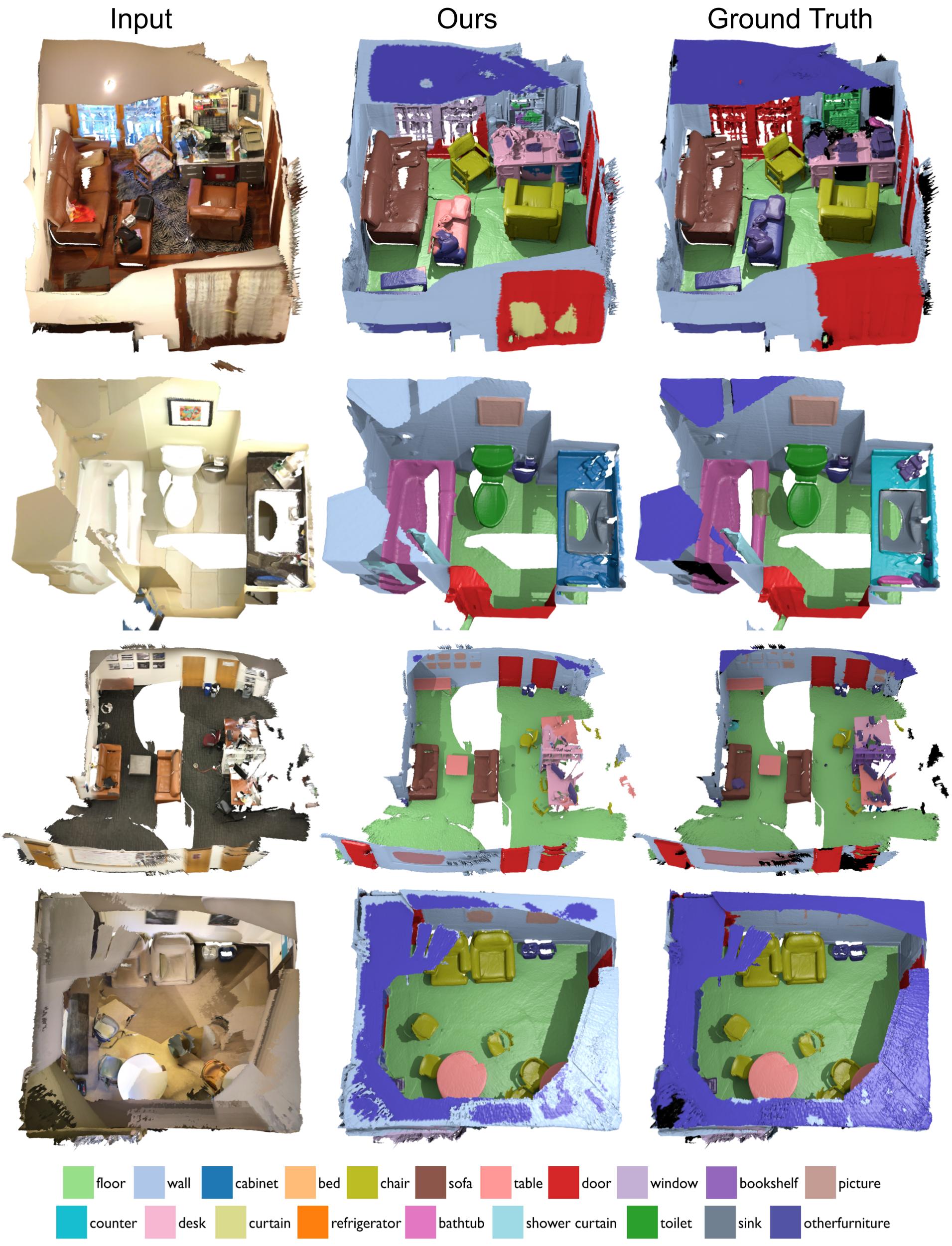}
\caption{Qualitative 3D semantic segmentation results on ScanNet validation set for scenes \texttt{scene0050\_00}, \texttt{scene0664\_01}, \texttt{scene0329\_02}, and \texttt{scene0616\_01} respectively.}
\label{fig:qualitative_scannet_val2}
\end{figure}

Qualitative results on ScanNet validation set are available in Figure~\ref{fig:qualitative_scannet_val1} and Figure~\ref{fig:qualitative_scannet_val2}. Also more detailed results with per class segmentation IoU scores are available in Table~\ref{tab:scannet_per_class_3d} and Table~\ref{tab:scannet_per_class_2d}.

\begin{table}
\scriptsize
\centering
\resizebox{\textwidth}{!}{%
\begin{tabular}{|c|c|cccccccccccccccccccc|}
\hline
% TABLE HEADER
\grayrowcolor Method & \rot{mean IoU} & \rot{bathtub}	&\rot{bed}	&\rot{bookshelf}	&\rot{cabinet}	&\rot{chair}	&\rot{counter}	&\rot{curtain}	&\rot{desk}	&\rot{door}	&\rot{floor}	&\rot{other furniture}	&\rot{picture} &\rot{refrigerator}	&\rot{shower curtain}	&\rot{sink}	&\rot{sofa}	&\rot{table}	&\rot{toilet}	&\rot{wall}	&\rot{window}  \\ 
\hline
PointNet~\cite{qi2017pointnet}              &55.7 &	73.5 &	66.1 &	68.6 &	49.1 &	74.4 &	39.2 &	53.9 &	45.1 &	37.5 &	94.6 &	37.6 &	20.5 &	40.3 &	35.6 &	55.3 &	64.3 &	49.7 &	82.4 &	75.6 &	51.5\\
PointConv~\cite{wu2019pointconv}            &66.6 &	78.1 &	75.9 &	69.9 &	64.4 &	82.2 &	47.5 &	77.9 &	56.4 &	50.4 &	95.3 &	42.8 &	20.3 &	58.6 &	75.4 &	66.1 &	75.3 &	58.8 &	90.2 &	81.3 &	64.2 \\
PointASNL~\cite{yan2020pointasnl}           &66.6 &	70.3 &	78.1 &	75.1 &	65.5 &	83.0 &	47.1 &	76.9 &	47.4 &  53.7 &	95.1 &	47.5 &	27.9 &	63.5 &	69.8 &	67.5 &	75.1 &	55.3 &	81.6 &	80.6 &	70.3\\
KPConv~\cite{thomas2019kpconv}              &68.4 & 84.7 &	75.8 &	78.4 &	64.7 &  81.4 &	47.3 &	77.2 &	60.5 &	59.4 &	93.5 &	45.0 &	18.1 &	58.7 &	80.5 &	69.0 &	78.5 &	61.4 &	88.2 &	81.9 &	63.2 \\
JSENet~\cite{jsenet}                        &69.9 &	88.1 &	76.2 &	82.1 &	66.7 &	80.0 &	52.2 &	79.2 &	61.3 &	60.7 &	93.5 &	49.2 &	20.5 &	57.6 &	85.3 &	69.1 &	75.8 &	65.2 &	87.2 &	82.8 &	64.9 \\
SparseConvNet~\cite{graham20183d}           &72.5 &	64.7 &	82.1 &	84.6 &	72.1 &	86.9 &	53.3 &	75.4 &	60.3 &	61.4 &	95.5 &	57.2 &	32.5 &	71.0 &	87.0 &	72.4 &	82.3 &	62.8 &	93.4 &	86.5 &	68.3  \\
MinkowskiNet~\cite{choy20194d}              &73.6 &	85.9 &	81.8 &	83.2 &	70.9 &	84.0 &	52.1 &	85.3 &  66.0 &	64.3 &	95.1 &	54.4 &	28.6 &	73.1 &	89.3 &	67.5 &	77.2 &	68.3 &	87.4 &  85.2 &	72.7 \\
\hline
\textbf{Ours} &\textbf{74.6} & 77.1 & 81.9 & 84.8 & 70.2 & 86.5 & 39.7 & 89.9 & 69.9 & 66.4 & 94.8 & 58.8 & 33.0 & 74.6 & 85.1 & 76.4 & 79.6 & 70.4 & 93.5 & 86.6 & 72.8  \\
\hline
\end{tabular}
}
\caption{Per class 3D semantic labeling results on the ScanNet test split.}
\label{tab:scannet_per_class_3d}
\end{table}

\begin{table}
\scriptsize
\centering
\resizebox{\textwidth}{!}{%
\begin{tabular}{|c|c|cccccccccccccccccccc|}
\hline
% TABLE HEADER
\grayrowcolor Method & \rot{mean IoU} & \rot{bathtub}	&\rot{bed}	&\rot{bookshelf}	&\rot{cabinet}	&\rot{chair}	&\rot{counter}	&\rot{curtain}	&\rot{desk}	&\rot{door}	&\rot{floor}	&\rot{other furniture}	&\rot{picture} &\rot{refrigerator}	&\rot{shower curtain}	&\rot{sink}	&\rot{sofa}	&\rot{table}	&\rot{toilet}	&\rot{wall}	&\rot{window}  \\ 
\hline
3DMV~\cite{dai20183dmv} &49.8 &	48.1 & 61.2 & 57.9 & 45.6 &	34.3 & 38.4 & 62.3 & 52.5 &	38.1 & 84.5 & 25.4 & 26.4 &	55.7 & 18.2 & 58.1 & 59.8 &	42.9 & 76.0 & 66.1 & 44.6 \\
\hline
AdapNet++~\cite{valada2019self} & 50.3 & 61.3 & 72.2 & 41.8 & 35.8 & 33.7 & 37.0 & 47.9 & 44.3 & 36.8 & 90.7 & 20.7 & 21.3 & 46.4 & 52.5 & 61.8 & 65.7 & 45.0 & 78.8 & 72.1 & 40.8 \\
\hline
SSMA~\cite{valada2019self} & 57.7 & 69.5 & 71.6 & 43.9 & 56.3 & 31.4 & 44.4 & 71.9 & 55.1 & 50.3 & 88.7 & 34.6 & 34.8 & 60.3 & 35.3 & 70.9 & 60.0 & 45.7 & 90.1 & 78.6 & 59.9 \\
\hline
\textbf{Ours} &\textbf{74.5} & 86.1 & 83.9 & 88.1 & 67.2 & 51.2 & 42.2 & 89.8 & 72.3 & 71.4 & 95.4 & 45.4 & 50.9 & 77.3 & 89.5 & 75.6 & 82.0 & 65.3 & 93.5 & 89.1 & 72.8  \\
\hline
\end{tabular}
}
\caption{Per class 2D semantic labeling results on the ScanNet test split.}
\label{tab:scannet_per_class_2d}
\end{table}
\end{document}